\documentclass[10pt,twocolumn,letterpaper]{article}

\usepackage{wacv}              

\usepackage{graphicx}
\usepackage{amsmath}
\usepackage{amssymb}
\usepackage{booktabs}
\usepackage{url}            
\usepackage{amsfonts}       
\usepackage{nicefrac}       
\usepackage{microtype}      
\usepackage{xcolor}         
\usepackage{multicol}
\usepackage{wrapfig}

\usepackage{bbm}
\usepackage{algorithm}
\usepackage{multirow}
\usepackage{diagbox}

\usepackage{tabularx}
\usepackage{arydshln}
\usepackage{caption}
\usepackage{subcaption}

\usepackage[pagebackref,breaklinks,colorlinks]{hyperref}

\usepackage[capitalize]{cleveref}
\crefname{section}{Sec.}{Secs.}
\Crefname{section}{Section}{Sections}
\Crefname{table}{Table}{Tables}
\crefname{table}{Tab.}{Tabs.}

\def\wacvPaperID{2121} 
\def\confName{WACV}
\def\confYear{2025}

\begin{document}

\title{Revisiting Machine Unlearning with Dimensional Alignment}

\author{
Seonguk Seo$^{1}$ \qquad  Dongwan Kim$^1$ \qquad Bohyung Han$^{1,2}$ \\
$^1$ECE \& $^{2}$IPAI, Seoul National University\\
   {\tt\small \{seonguk, dongwan123, bhhan\}@snu.ac.kr}
}
\maketitle

\begin{abstract}
Machine unlearning, an emerging research topic focusing on data privacy compliance, enables trained models to erase information learned from specific data.
While many existing methods indirectly address this issue by intentionally injecting incorrect supervision, they often result in drastic and unpredictable changes to decision boundaries and feature spaces, leading to training instability and undesired side effects. 
To address this challenge more fundamentally, we first analyze the changes in latent feature spaces between the original and retrained models, and observe that the feature representations of samples not included in training are closely aligned with the feature manifolds of previously seen samples. 
Building on this insight, we introduce a novel evaluation metric for machine unlearning, coined \textit{dimensional alignment}, which measures the alignment between the eigenspaces of the forget and retain sets.
We incorporate this metric as a regularizer loss to develop a robust and stable unlearning framework, which is further enhanced by a self-distillation loss and an alternating training scheme.
Our framework effectively eliminates information from the forget set while preserving knowledge from the retain set.
Finally, we identify critical flaws in existing evaluation metrics for machine unlearning and propose new tools that more accurately capture its fundamental objectives.

\end{abstract}




\section{Introduction}
\label{sec:intro}

Deep neural networks have demonstrated remarkable advances across various domains, achieving impressive performance by leveraging large-scale data.
However, despite their success, these models are susceptible to unintentionally memorizing training data~\cite{zhang2021understanding}, making them vulnerable to inference attacks that could compromise user privacy and expose sensitive information from the training data.
In response to growing privacy concerns, regulations such as General Data Protection Regulation (GDPR) in the European Union and the California Consumer Privacy Act (CCPA) grant individuals the ``right to be forgotten", enabling users to demand the deletion of their personal data by service providers.
This regulatory environment necessitates the concept of \textit{machine unlearning}, a process systematically removing the information about specific examples from trained models.
The primary goal of machine unlearning is to ensure that once data is removed from a model, the model behaves as it had never been trained on the data.
This concept is crucial for complying with privacy laws and maintaining ethical standards in machine learning applications.

The exact and straightforward solution for machine unlearning is to retrain the model from scratch, excluding the data requested for deletion.
While this approach ensures that models remain completely unaffected by the data to be forgotten, it is impractical due to the excessive computational costs and the need for access to the full training dataset.
To address this challenge, machine unlearning research has shifted towards developing faster approximate methods, where the goal is to finetune a trained model such that it becomes indistinguishable from one that has undergone exact unlearning.
A prominent theme for approximate unlearning approaches has been to intentionally inject incorrect supervisions for the samples to be forgotten~\cite{thudi2022unrolling, chundawat2023can, kurmanji2024towards, cha2024learning,graves2021amnesiac}, such as training with random labels or reversed gradients, which reformulates unlearning as \textit{mislearning}.
However, the goal of unlearning is not merely to make incorrect predictions for the examples to be forgotten, but to erase the information additionally learned from the forget set.
Furthermore, mislearning approaches often lead to over-forgetting, which degrades the model's overall performance and training stability.

To thoroughly explore the nature of unlearning and identify the necessary steps to achieve it, we begin by analyzing the behavior of latent feature representations in an incremental learning scenario, which mirrors the reverse process of unlearning.
When visualizing the feature representations of initially unseen samples, we first observe that they align with the feature manifolds of previously seen samples.
However, after the model is trained on the unseen samples, their feature representations shift to enhance discrimination.
This behavior suggests that, as the reverse process of incremental learning, unlearning should reposition the forget samples within the feature space of the retained samples.
In this context, we propose \textit{dimensional alignment}, a novel evaluation metric for machine unlearning that measures the alignment between the feature spaces of the forget and retain sets.
We employ this metric as a regularization loss term to achieve robust unlearning and introduce a simple yet effective self-distillation loss to further enhance stability.
Building on these ideas, we propose a comprehensive unlearning framework, termed Machine Unlearning with Dimensional Alignment (MUDA), which integrates an alternating training scheme with the proposed loss functions, to ensure robust unlearning and preserve knowledge from the retain set.

Finally, we address the limitation of current evaluation metrics for machine unlearning.
Most existing unlearning approaches typically adopt evaluation metrics based on final outputs, such as forget set accuracy or membership inference attack score.
However, since discriminative models produce low-dimensional score vectors that do not explicitly reveal sensitive information, these outputs alone might not be sufficient to confirm successful unlearning in classification tasks.
Our empirical observations show that existing evaluation metrics can be easily manipulated through trivial fine-tuning of the last linear layer.
Given that the primary goal of unlearning is to prevent information leakage from the samples to be forgotten, it is crucial to concentrate on the latent feature representations that carry semantic information.
To achieve this, we present a collection of evaluation metrics---dimensional alignment, linear probing, F1 score, and normalized mutual information--that together more accurately reflect the primary objectives of machine unlearning.

Our main contributions are summarized as follows.
\begin{itemize} 

    \item We propose a novel metric, coined \textit{dimensional alignment}, to analyze machine unlearning in the latent feature space, which measures the alignment between the feature spaces of the forget and retain sets.
    Notably, dimensional alignment also serves as an effective training objective for machine unlearning.

    \item We propose a self-distillation loss designed to ensure stable unlearning while minimizing negative impacts on the feature representations of the retain set.

    \item We propose a novel framework for machine unlearning, referred to as MUDA, which incorporates an alternating training scheme along with the dimensional alignment and self-distillation losses to ensure effective and stable unlearning.

    \item We highlight the shortcomings of current evaluation metrics and introduce new feature-level evaluation metrics that more accurately reflect the objectives of machine unlearning.

\end{itemize}

The rest of the paper is organized as follows.
We review the preliminaries in Section~\ref{sec:prelim}.
Section~\ref{sec:method} presents the proposed approach in the context of machine learning and Section~\ref{sec:verification} discusses the evaluation protocols.
We validate the effectiveness of our frameworks in Section~\ref{sec:exp} and review the prior works in Section~\ref{sec:related}.
Finally, we conclude our paper in Section~\ref{sec:conclusion}.


\section{Preliminaries}
\label{sec:prelim}

\subsection{Machine unlearning}
\label{sec:unlearning}

Let us consider a neural network model, $f(\cdot; \theta_o)$, parameterized by $\theta_o$, initially trained on a dataset $\mathcal{D} = \{ (x_i, y_i) \}^N_{i=1}$ which consists of $N$ pairs of the input data $x_i$ and its corresponding class label $y_i$. 
The goal of machine unlearning is to remove the influence of a forget set, $\mathcal{D}_f \subseteq \mathcal{D}$, from the original model, $\theta_o$, while preserving utility over the retain set, $\mathcal{D}_r = \mathcal{D} \setminus \mathcal{D}_f$.

An straightforward unlearning strategy refers to training a new model, $\theta_r$, only using the retain set, $\mathcal{D}_r$. 
Although this solution meets the condition for unlearning, it entails a huge computational burden especially when the training dataset is large or unlearning request happens frequently.
To alleviate this issue, \textit{approximate} unlearning approaches aim to derive an unlearned model $\theta_u$ from the original model $\theta_o$, where $\theta_u$ is statistically indistinguishable from the retrained model $\theta_r$.

\subsection{Setting}
\label{sec:setting}
Our work investigates machine unlearning under the context of image classification.
We specifically focus on \textit{class unlearning} and \textit{subclass unlearning}, where the forget set $\mathcal{D}_f$ consists of samples belonging to a specific class and subclass, respectively.\footnote{We refer to Section~\ref{sec:random_sample_unlearning} for random sample unlearning.}

There are no well-defined constraints on the amount of data used for machine unlearning. 
However, as mentioned in Section~\ref{sec:unlearning}, using the entire retain set $\mathcal{D}_r$ entails a large computational burden.
Therefore, we opt to use only a subset of $\mathcal{D}_r$, which we denote as $\mathcal{D}_r'$, to train the unlearned model.
$\mathcal{D}_r'$ is randomly sampled from $\mathcal{D}_r$ such that $|\mathcal{D}_r'| = |\mathcal{D}_f|$.


\section{Machine Unlearning with Dimensional Alignment (MUDA)}
\label{sec:method}

\subsection{Unlearning as a reverse process of incremental learning}
\label{sec:continual}

\begin{figure*}[t]
\centering
\includegraphics[width=0.8\linewidth]{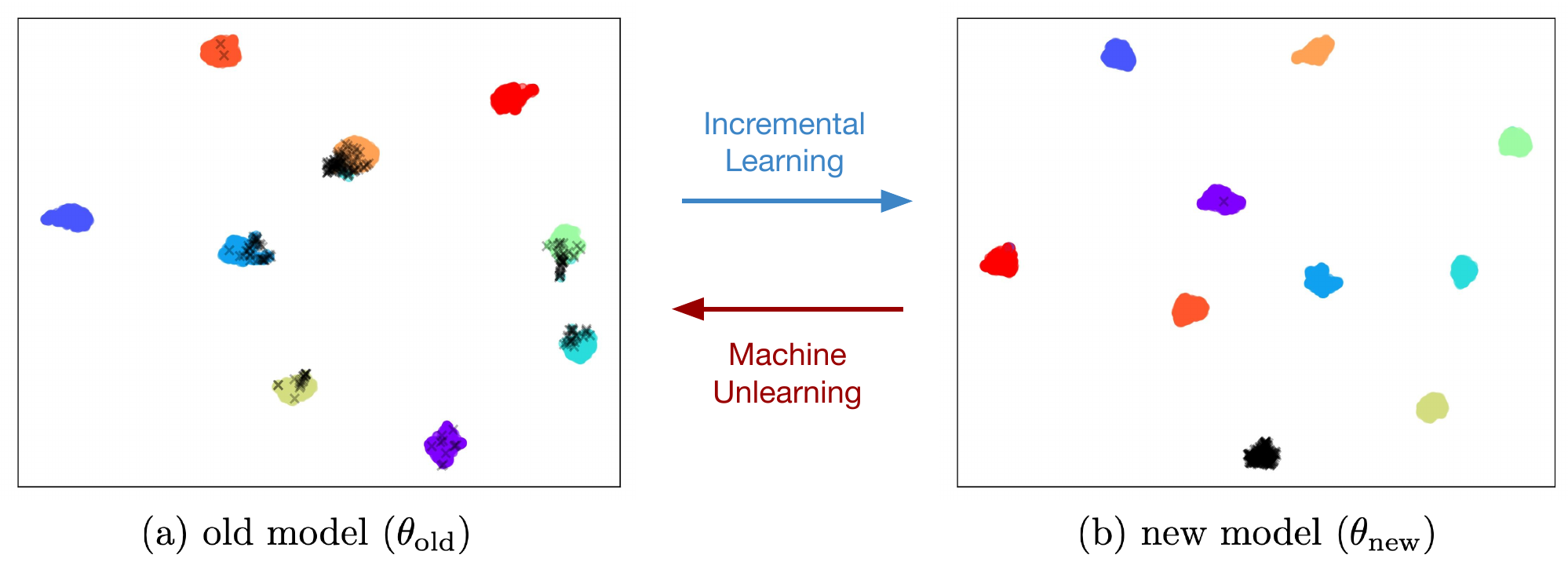}
\caption{UMAP visualization of CIFAR-10 train set under the incremental learning scenario, where old and new models are trained with $\mathcal{D}_r$ and $\mathcal{D}_r \cup \mathcal{D}_f$, respectively.
    Black markers indicate the feature representations of $\mathcal{D}_f$.
    }
    \label{fig:feature_umap}
\end{figure*}

One way to interpret machine unlearning is as a reverse process of incremental learning.
In the concept of incremental learning, a model $\theta_\text{old}$ is initially trained on an old dataset $\mathcal{D}_\text{old}$ and subsequently trained on a new dataset $\mathcal{D}_\text{new}$, where the goal is for the new model, $\theta_\text{new}$, to perform well on the combined dataset $\mathcal{D} = \mathcal{D}_\text{old} \cup \mathcal{D}_\text{new}$.
The parallels between machine unlearning and incremental learning are evident.

Building on this, we examine an incremental learning model to gain insights for machine unlearning.
We are particularly interested in observing how the feature representations shift from $\theta_\text{old}$ to $\theta_\text{new}$ to better understand how the reverse process (\textit{i.e.,} unlearning) should behave. 
To this end, we train a ResNet-18 model on the CIFAR-10 dataset under the incremental learning setting, where $\mathcal{D}_\text{old}$ includes samples from classes 1$\sim$9 and $\mathcal{D}_\text{new}$ consists of samples from class 10.
Figure~\ref{fig:feature_umap} illustrates a UMAP~\cite{mcinnes2018umap} visualization of feature representations generated by $\theta_\text{old}$ and $\theta_\text{new}$.
Each color represents a different class from the CIFAR-10 dataset, with the new class 10 samples distinctly marked by black crosses.

As depicted in Figure~\ref{fig:feature_umap}(a), feature representations of the new class 10 samples, which were not included in the training of $\theta_\text{old}$, are dispersed across the feature space of the seen classes.
These samples tend to gravitate towards the classes they share the most similarities with. 
Conversely, Figure~\ref{fig:feature_umap}(b) demonstrates the shift in representations after incorporating class 10 into the training data.
After training, these new samples shift away from the feature manifold of the old classes and cluster together.
This shift suggests that incremental learning adjusts the representations of the new samples by moving them to a new feature manifold, thereby enhancing their semantic clarity.
Thus, from this perspective, we can perceive machine unlearning as the process of projecting the feature representations of the forget samples back onto the feature manifold of the retain set.

\begin{table*}[!t]
\centering
\caption{Evaluation results for dimensional alignment, DA($\mathcal{D}_f|\mathcal{D}_r; \theta$), across various settings and datasets.
The results are averaged over five runs, each with varying forget (sub)class, for every configurations.
}
\label{tab:dimensional_alignment}
\setlength\tabcolsep{10pt}
\vspace{0.2cm}
\scalebox{0.95}{
\begin{tabular}{@{}llcccc@{}}
\toprule
& & \multicolumn{3}{c}{Class unlearning} & \multicolumn{1}{c}{Subclass unlearning} \\
\cmidrule(lr){3-5} \cmidrule(l){6-6} 
Method & Train set & CIFAR-10 & CIFAR-100 & Tiny-ImageNet & CIFAR-20 \\ 
\midrule
Original &$\mathcal{D}_r \cup \mathcal{D}_f$ &0.34~\footnotesize{$\pm$0.05} &0.50~\footnotesize{$\pm$0.06} &0.59~\footnotesize{$\pm$0.04} &0.48~\footnotesize{$\pm$0.09} \\
Retrained &$\mathcal{D}_r$ &0.79~\footnotesize{$\pm$0.04} &0.74~\footnotesize{$\pm$0.03} &0.73~\footnotesize{$\pm$0.04} &0.84~\footnotesize{$\pm$0.04} \\
\bottomrule
\end{tabular}
}
\end{table*}

\subsection{Dimensional alignment}
\label{sec:da}

To achieve the goal of projecting the feature representations of the forget samples onto the manifold of the retain set, we start by measuring the alignment between the two feature spaces.  
Let $\mathbf{F}_r \in \mathbb{R}^{C\times |\mathcal{D}_r|}$ and $\mathbf{F}_f \in \mathbb{R}^{C \times |\mathcal{D}_f|}$ denote the $C$-dimensional feature representations extracted by model $\theta$ for $\mathcal{D}_r$ and $\mathcal{D}_f$, respectively.
We compute the eigenvectors of covariance matrix for the retain set by singular value decomposition (SVD), \ie, $\mathbf{F}_r\mathbf{F}_r^T = \mathbf{U}_r \mathbf{\Sigma}_r \mathbf{U}_r^T$.
Among the $C$ eigenvectors of $\mathbf{U}_r$, we keep the $k$ eigenvectors corresponding to the top-$k$ largest eigenvalues, $\widehat{\mathbf{U}}_r = [\mathbf{u}_1, ... , \mathbf{u}_k]^T$, where $k$ is determined by the effective rank~\cite{roy2007effective} of the covariance matrix.
Then, we define the \textbf{dimensional alignment (DA)} as
\begin{align}
\label{eq:da}
 \text{DA}(\mathcal{D}_f | \mathcal{D}_r; \theta) := \lVert \mathbf{F}_f \mathbf{F}_f^T \widehat{\mathbf{U}}_r \widehat{\mathbf{U}}_r^T \rVert_F / \lVert \mathbf{F}_f \mathbf{F}_f^T \rVert_F,
\end{align}
where $\lVert \cdot \rVert_F$ denotes the Frobenius norm.

\begin{figure}[t!]
  \begin{center}
    \includegraphics[width=0.4\textwidth]{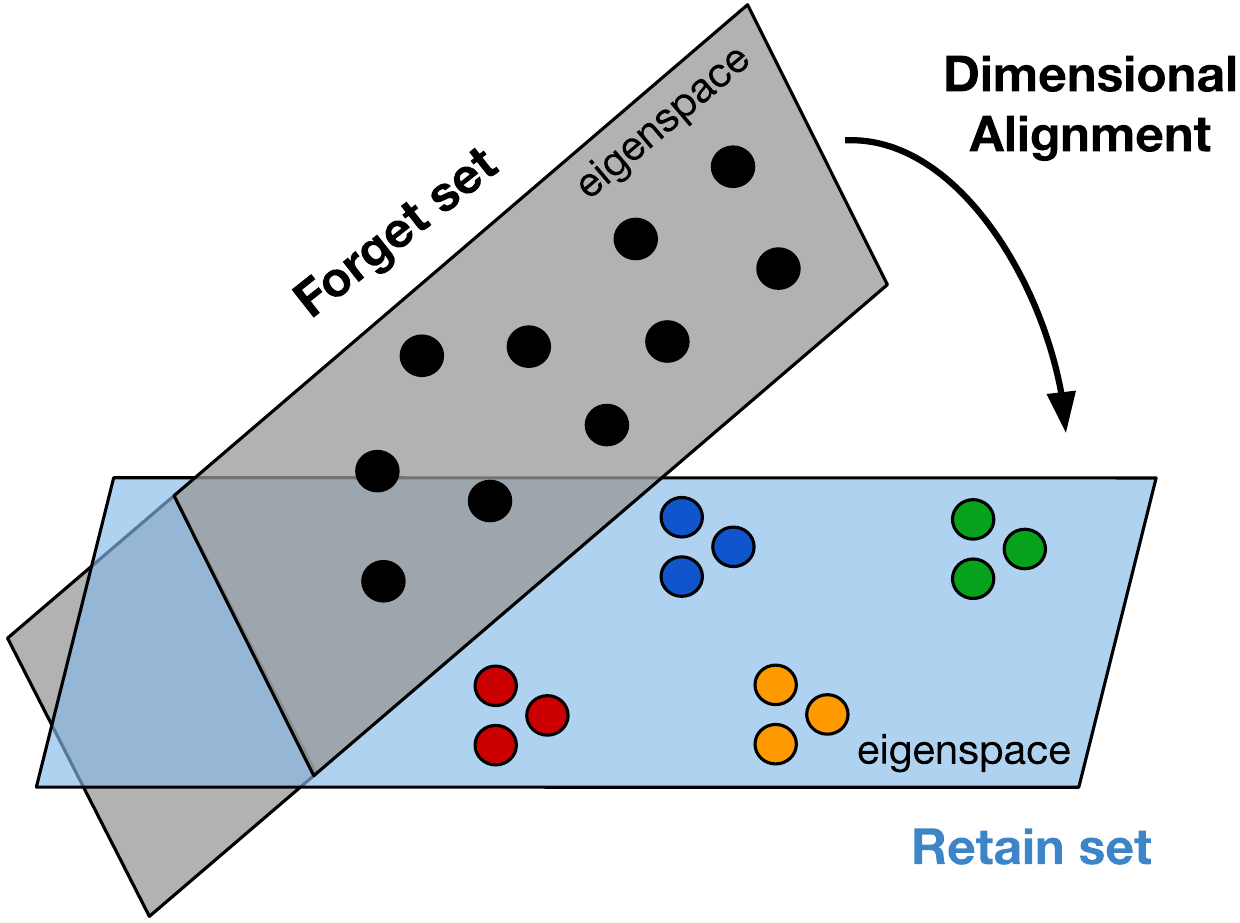}  
  \end{center}
  \caption{Conceptual visualization of dimensional alignment.}
  \label{fig:overview}
  \vspace{-3mm}
\end{figure}

Dimensional alignment (DA), as depicted in Figure~\ref{fig:overview}, measures how well the feature space of $\mathcal{D}_f$ aligns with the principal component subspace of $\mathcal{D}_r$. 
A higher DA value signifies a stronger alignment, indicating that the feature representations of $\mathcal{D}_f$ are well-aligned with the most significant dimensions of the feature representations of $\mathcal{D}_r$.  
As a sanity check, we measure DA($\mathcal{D}_f | \mathcal{D}_r; \theta$) across various unlearning settings and datasets.
The results in Table~\ref{tab:dimensional_alignment} demonstrates that $\theta_r$ consistently exhibits higher DA than $\theta_o$ in all cases, which is consistent with our observations from Section~\ref{sec:continual}. 
Thus, we posit that DA can serve as an effective metric for assessing the effectiveness of unlearning in feature representations.

Moreover, we can incorporate DA directly as a regularization term in the unlearning process, \ie, $\mathcal{L}_\text{DA} = -\text{DA}(\mathcal{D}_f|\mathcal{D}_r'; \theta)$\footnote{Note that we use $\mathcal{D}_r'$ for $\mathcal{L}_\text{DA}$ following the setting described in Section~\ref{sec:setting}}.
By applying a stop-gradient operation on $\mathbf{F}_r$ and updating only $\mathbf{F}_f$, we prevent distortion of the feature manifold of $\mathcal{D}_r'$ and ensure training stability.
This loss term facilitates unlearning by minimizing the information in $\mathcal{D}_f$ that is not already encoded by $\mathcal{D}_r'$.

\subsection{Self-distillation loss for stable projection onto retain feature manifold}
\label{sec:distill_loss}

Currently, most of unlearning methods utilize loss functions designed for \textit{mislearning}, such as training with reversed gradients~\cite{thudi2022unrolling, kurmanji2024towards}, but these approaches have significant drawbacks that can alter the decision boundaries and feature spaces in undesirable ways, leading to instability. 
Furthermore, prolonged training with reversed gradients can degrade the discriminability of other classes, compromising the model's overall utility.

To ensure effective unlearning with high stability, we propose a self-distillation~\cite{hinton2015distilling} loss on the forget set $\mathcal{D}_f$.
As illustrated in Figure~\ref{fig:feature_umap}(a), this self-distillation aims to redistribute the forget samples towards the retain classes.
Unlike $\mathcal{L}_\text{DA}$ which uses feature representations, the self-distillation is applied on the softmax outputs of the model, $f(x; \theta)$. 
In order to redistribute the forget samples to other classes, the distillation target must represent the output probability of all classes as if the forget class did not exist.
In practice, we can simply set the value of $f(x; \theta)$ corresponding to the forget class as 0 and renormalize to obtain our distillation target, $\hat{f}(x; \theta)$. 
Then, the self-distillation loss can be expressed as
\begin{align}
  \mathcal{L}_\text{SD} = \frac{1}{|\mathcal{D}_f|}\sum_{x \in \mathcal{D}_f} D_\text{KL}\Big(f\big(x; \theta\big) \Vert \hat{f}\big(x; \theta\big) \Big),
  \label{eq:sd}
\end{align}
where $D_\text{KL}(\cdot)$ is the KL-divergence loss.
Note that the distillation target, $\hat{f}(x; \theta)$, is dynamic and evolves throughout the training process. 
Consequently, $\mathcal{L}_\text{SD}$ seeks to establish an equilibrium between $f(x; \theta)$ and $\hat{f}(x; \theta)$, offering a highly stable objective compared to training on reversed gradients, and thereby eliminating the need for early stopping to ensure strong model performance.

\subsection{Overall framework}

Our overall loss function used for unlearning is

\begin{align}
 \mathcal{L} = \underbrace{ \alpha \cdot \mathcal{L}_\text{DA} + \beta \cdot  \mathcal{L}_\text{SD} \vphantom{\sum_{(x, y) \in \mathcal{D}_r}}}_{\text{forget phase}}
 + \underbrace{ \frac{1}{|\mathcal{D}_r'|} \sum_{(x, y) \in \mathcal{D}_r'} \ell(f(x;\theta), y) }_{\text{recover phase}} , 
\end{align}
where $\alpha$ and $\beta$ are hyperparameters to balance the corresponding loss terms and $\ell(\cdot, \cdot)$ denotes the cross-entropy loss function.
The first two terms aim to mitigate the influence of $\mathcal{D}_f$ whereas the last term serves to preserve the discriminability on $\mathcal{D}_r'$.
However, optimizing all terms simultaneously may cause some destructive interference, thereby diminishing the overall effectiveness.
To address this, we use an alternating training scheme that switches between \textit{forget} and \textit{recover} phases after each epoch.
In the forget phase, we remove the information of $\mathcal{D}_f$ using $\mathcal{L}_\text{DA}$ and $\mathcal{L}_\text{SD}$.
Subsequently, in the recover phase, we reinstate the knowledge of $\mathcal{D}_r'$ that might have been compromised.
This process is repeated until convergence.
We find that this alternating training scheme, combined with the proposed loss functions, effectively eliminates the knowledge of $\mathcal{D}_f$ while minimizing the loss of information on $\mathcal{D}_r'$.


\section{Verification of Machine Unlearning}
\label{sec:verification}

\begin{table*}[t]
\centering
\caption{Test results for existing evaluation metrics, averaging over five different configurations. Trivially finetuned models, applied only to the linear classifiers, achieve the desirable unlearning results across all datasets, despite not being actually unlearned.}
\label{tab:existing_metrics}
\setlength\tabcolsep{10pt}
\scalebox{0.88}{
\begin{tabular}{lcccccc}
\toprule
& \multicolumn{2}{c}{CIFAR-10} & \multicolumn{2}{c}{CIFAR-100} & \multicolumn{2}{c}{Tiny-ImageNet} \\
\cmidrule(lr){2-3}  \cmidrule(lr){4-5} \cmidrule(lr){6-7} 
 Method &  Acc($\mathcal{D}_f$) & MIA &  Acc($\mathcal{D}_f$) & MIA &  Acc($\mathcal{D}_f$) & MIA \\
\midrule
Original & 92.9 & 0.91 & 76.8 & 0.91 & 51.2 & 0.89 \\
Retrained & ~~0.0 & 0.37 & ~~0.0 & 0.18 & ~~0.0 & 0.14 \\
FT (classifier only) & ~~0.0 & 0.00 & ~~0.0 & 0.01 & ~~0.0 & 0.18 \\
\bottomrule
\end{tabular}
}
\label{tab:results}
\end{table*}

\subsection{Limitations of existing metrics}

As mentioned in Section~\ref{sec:unlearning}, the primary goal of approximate unlearning methods is to derive an unlearned model, $\theta_u$, that is statistically indistinguishable from the retrained model, $\theta_r$.
To determine whether $\theta_u$ is statistically similar to $\theta_r$, previous works often rely on the following two metrics:
\begin{itemize}
    \item \textbf{Forget set accuracy:} 
    The accuracy on the forget set given model parameters $\theta$, Acc($\mathcal{D}_f | \theta$), is often used to evaluate how well the forget samples are unlearned.
    In the context of class unlearning, the forget set accuracy of the retrained model, Acc($\mathcal{D}_f ; \theta_r$), is ideally $0\%$; similarly, Acc($\mathcal{D}_f ; \theta_u$) should also be $0\%$.
    \item \textbf{MIA success rate:} 
    Membership inference attack~\cite{shokri2017membership} (MIA) is a type of privacy attack where an adversary attempts to predict whether a particular data sample was used to train a machine learning model.
    Ideally, the membership status of $\mathcal{D}_f$ should be predicted as non-training samples, leading in a lower MIA success rate.
\end{itemize}

However, we find that optimal scores for both metrics can be achieved by simply finetuning (FT) the classifier of $\theta_o$ (a single linear layer) on a small subset of $\mathcal{D}_r$. 
Table~\ref{tab:existing_metrics} demonstrates this by evaluating the original, retrained, and trivially finetuned models on three datasets using the Acc($\mathcal{D}_f | \theta$) and MIA metrics. 
As shown, the finetuned model achieves $0\%$ accuracy on the forget set across all datasets and even outperforms the retrained model in terms of MIA.
These results suggest that both the forget set accuracy and MIA success rate metrics may not accurately reflect model unlearning performance, as they can be easily manipulated without any actual unlearning\footnote{We also describe a few other methods to exploit Acc($\mathcal{D}_f | \theta$) and MIA in Section~\ref{supp:metric_limitations} of the Appendix.}. 
Therefore, it is essential to reconsider the fundamental purpose of model unlearning and develop metrics that better represents unlearning efficacy.

\subsection{Evaluation metrics with semantic information}
\label{sec:measure_sem_info}

The primary purpose of unlearning is to prevent information leakage, particularly relating to privacy breaches.
For generative models, the outputs embody semantic information, often including sensitive details such as images or texts.
Therefore, it is crucial to focus on the outputs of these models to prevent the generation of sensitive content.
In contrast, discriminative models output low-dimensional score vectors, which typically do not explicitly reveal any sensitive information and can be easily manipulated, as also shown in Table~\ref{tab:existing_metrics}.
Consequently, to better evaluate machine unlearning in discriminative models, we argue that validation metrics should focus on the feature representations, which encode sensitive semantic information, rather than on the outputs.
To achieve this, we utilize linear probing (LP), F1 score, and normalized mutual information (NMI) metrics to quantify the level of semantic information pertaining to $\mathcal{D}_f$ that is encoded in the unlearned model.

\paragraph{Linear Probing} 
Linear probing (LP) has been extensively used to evaluate the quality of feature representations extracted by pretrained models~\cite{he2022masked,caron2021emerging,alain2016understanding}, and has also been used to analyze the degree of stability and plasticity changes in recent continual learning methods~\cite{kim2023stability}.
Consequently, it is equally feasible to employ LP to evaluate the effectiveness of information elimination in the context of unlearning.

\paragraph{F1 and NMI} 
In addressing privacy leakage, we also adopt the F1 score and normalized mutual information (NMI) index.
Both the F1 and NMI metrics assess the identifiability of $\mathcal{D}_f$ through clustering based on feature representations, with higher values suggesting $\mathcal{D}_f$ is more easily identifiable. 
More details regarding the implementation of F1 and NMI are provided in the Appendix.


\section{Experiment}
\label{sec:exp}

\subsection{Experimental setup}

\paragraph{Datasets and baselines}

We conduct experiments on the standard benchmarks for machine unlearning: CIFAR-10, CIFAR-100~\cite{krizhevsky2009learning}, and Tiny-ImageNet~\cite{le2015tiny}.
We compare our framework, MUDA,  with existing approximate unlearning approaches, which include Finetuning (FT)~\cite{warnecke2021machine}, NegGrad~\cite{thudi2022unrolling}, SCRUB~\cite{kurmanji2024towards}, Fisher Forgetting~\cite{golatkar2020eternal}, Exact Unlearning-$k$~\cite{goel2022evaluating}, Catastrophic Forgetting-$k$~\cite{goel2022evaluating}.
We additionally employ NegGrad+FT, which alternates each epoch between maximizing the classification loss on $\mathcal{D}_f$ and minimizing it on $\mathcal{D}_r$.
To reproduce the compared approaches, we primarily follow the settings from their original papers, adjusting the parameters only when it leads to improved performance.


\begin{table*}[t]
\centering
\caption{Class unlearning results on the CIFAR-10 dataset averaging over five different configurations.
Values in parentheses indicate the absolute difference from the ``Retrained'' setting, and those with the smallest absolute difference are bolded.
}
\label{tab:cifar10_class}
\setlength\tabcolsep{9pt}
\scalebox{0.88}{
\hspace{-3mm}
\begin{tabular}{ll|ccccc}
\toprule
Method & Train set & DA($\mathcal{D}_f | \mathcal{D}_r$) & LP($\mathcal{D}_f$) & LP($\mathcal{D}_r$) & F1 & NMI \\
\midrule
Original & - & 0.34 & 92.9 & 92.5 & 0.99 & 0.96 \\
Retrained & $\mathcal{D}_r$ & 0.79 & 65.4 & 92.1 & 0.54 & 0.31 \\
\midrule
FT & $\mathcal{D}_r'$ & 0.60 (0.19) & 81.8 (16.4) & 90.6 (1.5) & 0.72 (0.18) & 0.50 (0.19) \\
NegGrad & $\mathcal{D}_f$ & 0.55 (0.24) & 66.8 (1.4)~~ & 90.2 (1.9) & 0.51 (0.03) & 0.23 (0.08) \\
NegGrad+FT & $\mathcal{D}_r' \cup \mathcal{D}_f$ & 0.67 (0.12) & 75.8 (10.4) & 91.6 (0.5) & 0.56 (0.02) & 0.35 (0.04) \\
Fisher & $\mathcal{D}_r'$ & 0.37 (0.42) & 88.5 (23.1) & 90.2 (1.9) & 0.97 (0.43) & 0.89 (0.58) \\
SCRUB & $\mathcal{D}_r' \cup \mathcal{D}_f$ & 0.41 (0.38) & 74.7 (9.3)~~ & 92.0 (0.2) & 0.76 (0.22) & 0.59 (0.28) \\
EU-$k$ & $\mathcal{D}_r'$ & 0.73 (0.06) & 68.1 (2.7)~~ & 90.7 (1.4) & 0.73 (0.19) & 0.46 (0.14) \\
CF-$k$ & $\mathcal{D}_r'$ & 0.60 (0.19) & 81.3 (15.9) & \textbf{92.1 (0.0)} & 0.66 (0.12) & 0.43 (0.12) \\
MUDA (Ours) & $\mathcal{D}_r' \cup \mathcal{D}_f$ & \textbf{0.79 (0.00)} & \textbf{66.4 (1.0)}~~ & 92.3 (0.2) & \textbf{0.54 (0.00)} & \textbf{0.31 (0.00)} \\

\bottomrule
\end{tabular}
}
\end{table*}

\paragraph{Implementation details}

We adopt a ResNet-18~\cite{he2016deep} as the backbone network, where we replace the batch normalization with the group normalization~\cite{wu2018group}.
We train $\theta_o$ from scratch without pretraining using an SGD optimizer with a learning rate of 0.1, an exponential decay of 0.998, a weight decay of 0.001, and no momentum.
For unlearning, we use a learning rate of $1\times 10^{-3}$ over 200 iterations, setting $\alpha=0.1$ and $\beta=0.01$ unless specified otherwise.
Our framework is implemented using PyTorch~\cite{paszke2019pytorch} and experimented on NVIDIA RTX A5000 GPUs.
Please refer to Section~\ref{supp:implementation} in the Appendix for further implementation details.

\paragraph{Experimental configurations}
All experimental results are averaged over five different runs, each with a distinct construction of $\mathcal{D}_f$.
In the class unlearning setting, we construct $\mathcal{D}_f$ by choosing the forget class from classes \{1, 3, 5, 7, 9\} for CIFAR-10, classes \{1, 21, 41, 61, 81\} for CIFAR-100, and classes \{1, 41, 81, 121, 161\} for Tiny-ImageNet.
 In the subclass unlearning setting, we follow the approach in prior work~\cite{foster2024fast} and select five different subclasses\footnote{\textit{baby}, \textit{lamp}, \textit{mushroom}, \textit{rocket}, \textit{sea}} to be forgotten in CIFAR-20.
As detailed in Section~\ref{sec:setting}, for all approximate unlearning algorithms, we assume the availability of only a fixed subset of $\mathcal{D}_r$, referred to as $\mathcal{D}_r'$, during the unlearning process. 
To ensure the practicality of these algorithms, we set the size of $\mathcal{D}_r'$ to be equal to the size of $\mathcal{D}_f$.

\paragraph{Evaluation metrics}
We evaluate our method with the evaluation metrics proposed in Sections~\ref{sec:da} and~\ref{sec:measure_sem_info}, including: 1) dimensional alignment, DA($\mathcal{D}_f|\mathcal{D}_r$), 2) linear probing metrics, LP($\mathcal{D}_f$) and LP($\mathcal{D}_r$)\footnote{Note that we evaluate the corresponding test set of $\mathcal{D}_f$ and $\mathcal{D}_r$ for the LP($\cdot$) metric.}, 3) F1 score, and 4) NMI score.
Higher values are preferred for the DA($\mathcal{D}_f|\mathcal{D}_r$) and LP($\mathcal{D}_r$) metrics, whereas lower values are preferred for the other metrics.
Please refer to our supplementary document for the detailed descriptions.
We also report results using existing unlearning evaluation metrics, such as Acc($\mathcal{D}_f$), Acc($\mathcal{D}_r$), and the MIA score, in the Appendix.


\begin{table*}[t]
\centering
\caption{Class unlearning results on the Tiny-ImageNet dataset averaging over five different configurations.
Values in parentheses indicate the absolute difference from the ``Retrained'' setting, and those with the smallest absolute difference are bolded.
}
\label{tab:tiny_class}
\setlength\tabcolsep{9pt}
\scalebox{0.88}{
\hspace{-3mm}
\begin{tabular}{ll|ccccc}
\toprule
Method & Train set & DA($\mathcal{D}_f | \mathcal{D}_r$) & LP($\mathcal{D}_f$) & LP($\mathcal{D}_r$) & F1 & NMI \\
\midrule
Original & - & 0.59 & 47.6 & 58.0 & 0.96 & 0.92 \\
Retrained & $\mathcal{D}_r$ & 0.73 & 24.0 & 58.0 & 0.19 & 0.09 \\
\midrule
FT & $\mathcal{D}_r'$ & 0.60 (0.12) & 45.6 (21.6) & 56.2 (1.7)~~ & 0.66 (0.47) & 0.55 (0.46) \\
NegGrad & $\mathcal{D}_f$ & \textbf{0.74 (0.01)} & 29.6 (5.6)~~ & 55.1 (2.9)~~ & 0.22 (0.03) & 0.12 (0.03) \\
NegGrad+FT & $\mathcal{D}_r' \cup \mathcal{D}_f$ & 0.64 (0.09) & 37.2 (13.2) & 56.0 (2.0)~~ & 0.52 (0.33) & 0.38 (0.28) \\
Fisher & $\mathcal{D}_r'$ & 0.66 (0.07) & 34.8 (10.8) & 47.0 (11.0) & 0.39 (0.20) & 0.25 (0.16) \\
SCRUB & $\mathcal{D}_r' \cup \mathcal{D}_f$ & 0.64 (0.09) & 35.6 (11.6) & 55.8 (2.2)~~ & 0.52 (0.33) & 0.40 (0.30) \\
EU-$k$ & $\mathcal{D}_r'$ & 0.77 (0.05) & 12.8 (11.2) & 19.1 (38.9) & 0.11 (0.07) & 0.04 (0.05) \\
CF-$k$ & $\mathcal{D}_r'$ & 0.63 (0.10) & 40.8 (16.8) & 52.7 (5.3)~~ & 0.48 (0.29) & 0.33 (0.24) \\
MUDA (Ours) & $\mathcal{D}_r' \cup \mathcal{D}_f$ & 0.70 (0.03) & \textbf{26.0 (2.0)}~~ & \textbf{57.4 (0.6)}~~ & \textbf{0.21 (0.02)} & \textbf{0.10 (0.01)} \\
\bottomrule
\end{tabular}
}
\end{table*}

\subsection{Main results}

\paragraph{CIFAR-10}
The results for CIFAR-10 are presented in Table~\ref{tab:cifar10_class}. 
Judging by the drop in $\text{LP}(\mathcal{D}_f)$ from 92.9\% (original) to 66.4\%, our framework effectively eliminates information regarding $\mathcal{D}_f$, all the while maintaining discriminability on $\mathcal{D}_r$, as indicated by maintaining a high LP($\mathcal{D}_r$) of 92.3\%.
Furthermore, the DA, F1, and NMI metrics for our unlearned model is nearly identical to those of the retrained model, suggesting that the structure of the feature space of the two models are extremely well aligned. 
Among the baseline methods, NegGrad and EU-$k$ seem to successfully remove the knowledge of $\mathcal{D}_f$, but this comes at the cost of a 2\%p performance drop on LP($\mathcal{D}_r$).
On the other hand, SCRUB and CF-$k$ do not degrade performance on $\mathcal{D}_r$ but show limited effectiveness in unlearning.

\paragraph{Tiny-ImageNet and CIFAR-100}
The results for Tiny-ImageNet and CIFAR-100, as shown in Tables~\ref{tab:tiny_class} and~\ref{tab:cifar100_class_full} respectively, mostly mirror those observed with the CIFAR-10 dataset, where our framework remains effective according to all metrics.
Notably, EU-$k$ experiences a significant drop in performance in terms of LP($\mathcal{D}_r$). 
This decline is due to the algorithm's need to retrain the last few layers of the model from scratch, a process that is impractical in real-world scenarios with only a limited subset of training data available.


\begin{table*}[t]
\centering
\caption{Subclass unlearning results on the CIFAR-20 dataset averaging over five different configurations.
Values in parentheses indicate the absolute difference from the ``Retrained'' setting, and those with the smallest absolute difference are bolded.
}
\label{tab:cifar20_class}
\setlength\tabcolsep{9pt}
\scalebox{0.88}{
\hspace{-3mm}
\begin{tabular}{llccccc}
\toprule
Method  & Train set  & DA($\mathcal{D}_f | \mathcal{D}_r$) & LP$^\text{sub}$($\mathcal{D}_f$) & LP($\mathcal{D}_r$) & F1 & NMI \\
\midrule
Original & - & 0.48 & 60.6 & 81.7 & 0.34 & 0.25 \\
Retrained & $\mathcal{D}_r$ & 0.84 & 49.2 & 81.7 & 0.19 & 0.10 \\
\midrule
FT & $\mathcal{D}_r'$ & 0.74 (0.10) & 54.4 (5.2) & 81.6 (0.2) & 0.23 (0.03) & 0.13 (0.03) \\
NegGrad & $\mathcal{D}_f$ & 0.51 (0.33) & 53.4 (4.2) & 80.6 (1.1) & 0.33 (0.14) & 0.23 (0.13) \\
NegGrad+FT & $\mathcal{D}_r' \cup \mathcal{D}_f$ & 0.53 (0.31) & 57.6 (8.4) & 81.7 (0.0) & 0.34 (0.15) & 0.25 (0.15) \\
Fisher & $\mathcal{D}_r'$ & 0.68 (0.16) & 40.8 (8.4) & 75.7 (6.0) & 0.31 (0.12) & 0.20 (0.10) \\
SCRUB & $\mathcal{D}_r' \cup \mathcal{D}_f$ & 0.57 (0.28) & 56.6 (7.4) & \textbf{81.7 (0.0)} & 0.34 (0.14) & 0.23 (0.14) \\
EU-$k$ & $\mathcal{D}_r'$ & 0.76 (0.08) & 45.4 (3.8) & ~~53.4 (28.4) & \textbf{0.17 (0.02)} & \textbf{0.08 (0.02)} \\
CF-$k$ & $\mathcal{D}_r'$ & 0.50 (0.34) & 51.6 (2.4) & 81.6 (0.2) & 0.31 (0.12) & 0.23 (0.13) \\
MUDA (Ours) & $\mathcal{D}_r' \cup \mathcal{D}_f$ & \textbf{0.78 (0.07)} & \textbf{48.8 (0.4)} & 81.5 (0.3) & 0.15 (0.04) & 0.06 (0.04) \\
\bottomrule
\end{tabular}
}
\end{table*}

\paragraph{Overall}
Since the goal of machine unlearning is for the unlearned model to be statistically similar to the retrained model, we use the retrained model's score as a reference point for all metrics. 
Achieving a low difference between the unlearned and retrained models is ideal. 
Across all datasets and metrics, our framework consistently shows the greatest similarity to the retrained model, demonstrating its effectiveness.
We also highlight that our dimensional alignment (DA) metric consistently correlates well with other measurements across all algorithms and experimental settings.

\subsection{Analysis}

\label{sec:analysis}

\paragraph{Subclass unlearning}

We validate the compared algorithms under a subclass unlearning scenario, where $\mathcal{D}_f$ consists of samples belonging to a specific subclass.
We employ the CIFAR-20 dataset, which is a variant of the CIFAR-100 dataset that groups 100 classes into 20 coarser-grained superclasses based on semantic similarity.
For the subclass unlearning, we follow the approach in prior work~\cite{foster2024fast} and select five different subclasses\footnote{\textit{baby}, \textit{lamp}, \textit{mushroom}, \textit{rocket}, \textit{sea}} to be forgotten.
For evaluation, we primarily use the same evaluation protocol with class unlearning, and additionally adopt $\text{LP}_\text{sub}(\mathcal{D}_f)$, where a linear probing classifier is trained for subclass prediction, to assess the semantic information of $\mathcal{D}_f$ specifically.
Table~\ref{tab:cifar20_class} presents that our framework successfully eliminate the knowledge of subclass information, while maintaining the performance on the original task.

\paragraph{Ablation study}

We perform the ablative experiments on the CIFAR-10 dataset to analyze the effectiveness of the proposed loss functions.
Table~\ref{tab:ablation} presents the results when only the dimensional alignment and self-distillation losses, respectively, are employed in our framework.
The results show that each loss term plays a crucial role to achieve machine unlearning.

\begin{table}[t]
\centering
\caption{Ablative results of our framework on CIFAR-10 dataset.
For each metric, the smallest absolute difference from the retrained setting are bolded.
}
\label{tab:ablation}
\setlength\tabcolsep{9pt}
\scalebox{0.88}{
\hspace{-3mm}
\begin{tabular}{cc|ccc}
\toprule
$\mathcal{L}_\text{DA}$ & $\mathcal{L}_\text{SD}$ & DA($\mathcal{D}_f | \mathcal{D}_r$) & LP($\mathcal{D}_f$) & LP($\mathcal{D}_r$) \\
\midrule
\multicolumn{2}{c|}{Retrained} & 0.79 & 65.4 & 92.1 \\
\hline
& & 0.34 & 92.9 & {92.5} \\
\checkmark & & 0.76 & 69.6 & 91.3  \\
& \checkmark & 0.67 & 71.1 & \textbf{92.3} \\
\checkmark & \checkmark & \textbf{0.79} & \textbf{66.4} & \textbf{92.3} \\
\bottomrule
\end{tabular}
}
\end{table}


\begin{figure}[!t]
\captionof{table}{Unlearning results on a defending against backdoor attacks, each averaging over five different configurations.
The smallest absolute difference compared to the retrained model is highlighted.
}
\vspace{5pt}
\centering
\setlength\tabcolsep{8pt}
\scalebox{0.88}{
\begin{tabular}{lccc}
\toprule
Method &  ASR$\downarrow$ & Acc($\mathcal{D}_\text{clean}$)$\uparrow$ & DA \\
\midrule
Original & 99.52 & 91.00 & 0.54 \\
Retrained & ~~7.41 & 92.35 & 0.97 \\
\midrule
FT & 11.53 & 88.96 & 0.91 \\
NegGrad & 12.08 & 89.59 & 0.93 \\
NegGrad+FT & 24.49 & 90.71 & 0.89 \\
Fisher & 99.46 & 84.91 & 0.41 \\
SCRUB & 57.86 & 87.79 & 0.66 \\
EU-$k$ & 58.39 & 80.69 & 0.94 \\
CF-$k$ & 72.36 & 90.10 & 0.72 \\
MUDA (Ours) & ~~\textbf{8.29} & \textbf{91.21} & \textbf{0.96} \\
\bottomrule
\label{tab:cifar10_backdoor}
\end{tabular}
}
 \vspace{-2mm}
\label{tab:cifar10_backdoor}
\end{figure}%
%
%
\begin{figure}
\centering
\includegraphics[width=0.47\textwidth]{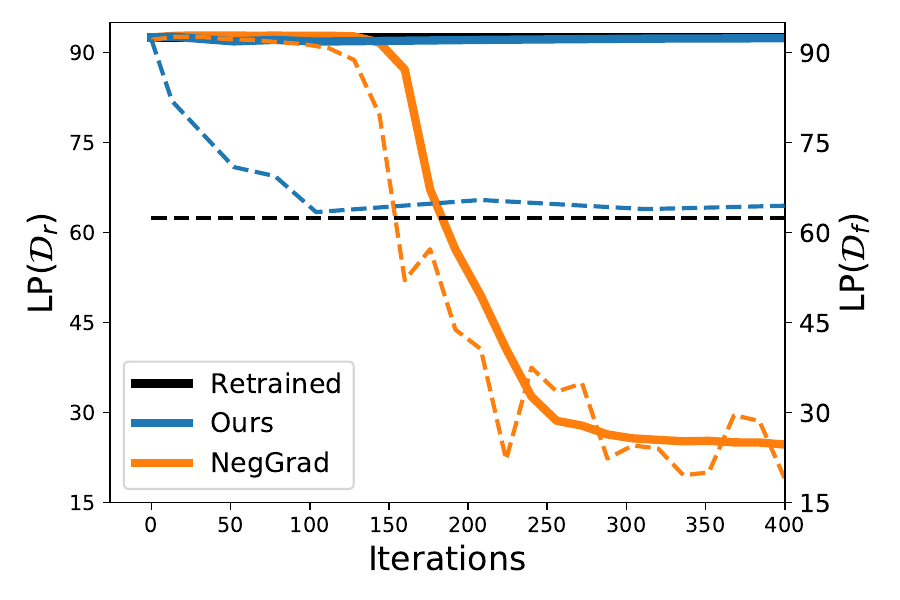}  
\captionof{figure}{
Visualizing the training stability.
Solid and dashed lines denote the results of LP($\mathcal{D}_r$) and LP($\mathcal{D}_f$).
Compared to NegGrad, which requires well-timed early stopping, our framework converges to a stable point.  
}
\label{fig:abl_stability}
\end{figure}

\paragraph{Defending against backdoor attack}

\begin{figure}[!t]
\hspace{-2mm}
    \centering
    \begin{subfigure}[b]{0.22\textwidth}  
        \centering
        \includegraphics[width=\textwidth]{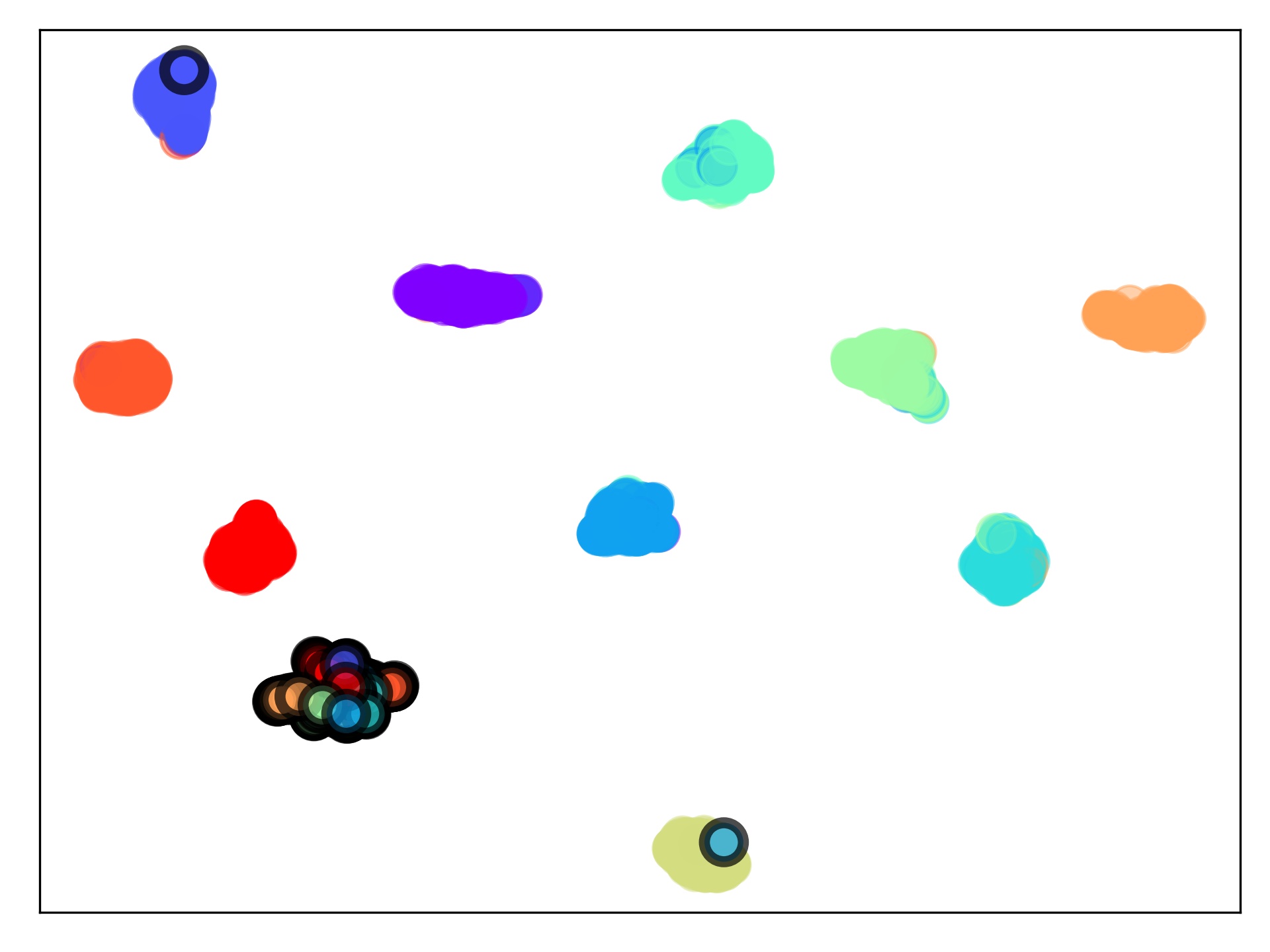} 
        \caption{Original model}
        \label{fig:umap_old}
    \end{subfigure}
    \begin{subfigure}[b]{0.22\textwidth} 
        \centering
        \includegraphics[width=\textwidth]{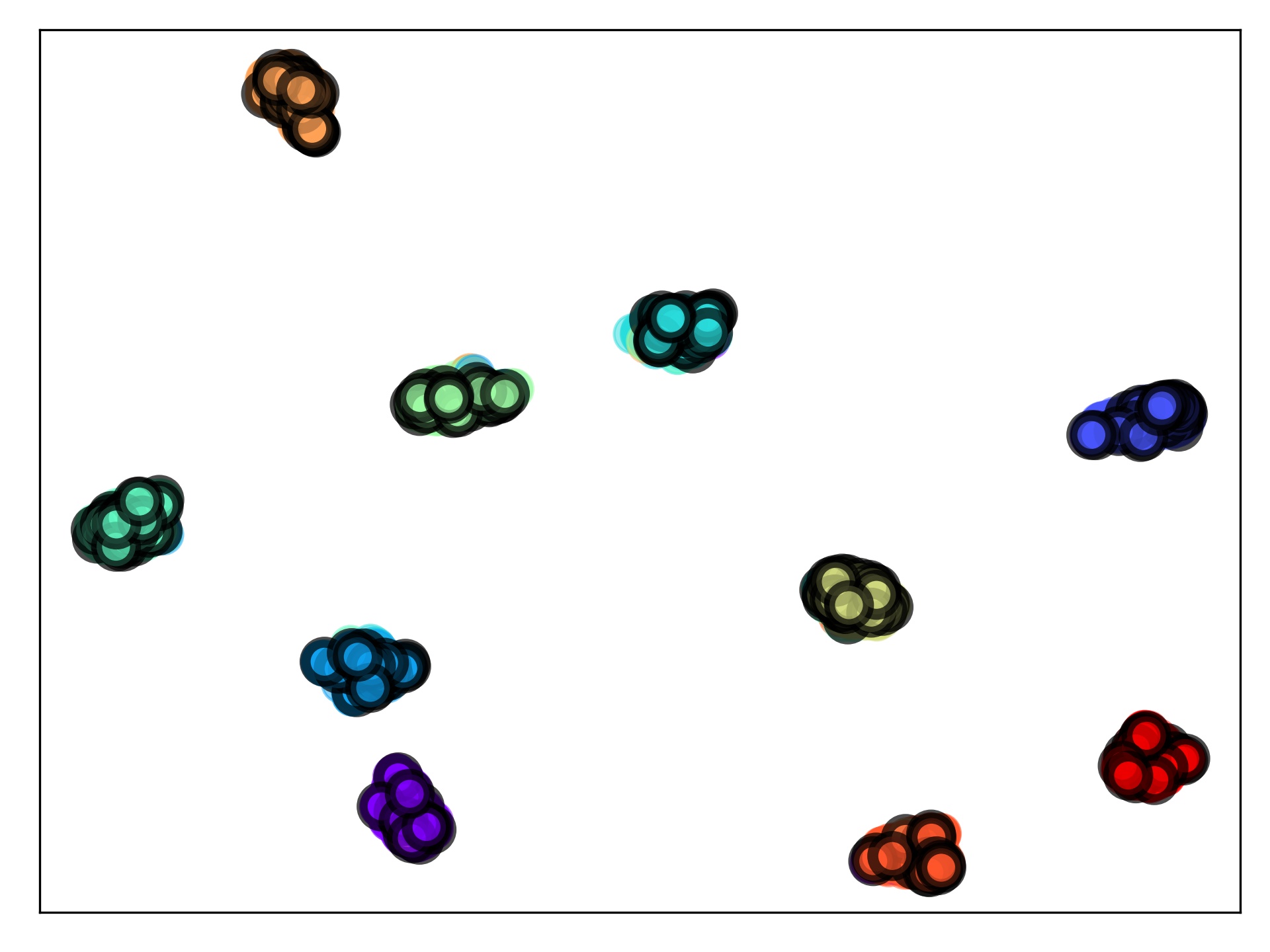} 
        \caption{Ours}
        \label{fig:umap_old}
    \end{subfigure}
    \vspace{2mm}
    \caption{UMAP visualization of CIFAR-10 train set under a backdoor attack scenario, (a) before unlearning and (b) after unlearning with our framework.
    Data points with black edges indicate the forget samples, which are poisoned by a backdoor trigger.
    }
    \label{fig:umap_backdoor}
\end{figure}

We evaluate our framework under a scenario where machine unlearning is employed as a means of defense against a backdoor attack~\cite{gu2017badnets, liu2024model}.
A backdoor attack is typically carried out by poisoning the training examples with triggers, \eg, a black patch that is associated with incorrect target labels.
When the trained model encounters inputs with these triggers, it incorrectly classifies them to the attacker's intended label, despite behaving normally on other inputs.
Our objective is to mitigate the effect of the backdoor trigger on model prediction by unlearning the poisoned samples.
For evaluation, we measure the backdoor attack success rate (ASR) and the accuracy on the clean test set, Acc($\mathcal{D}_\text{clean}$), as well as the dimensional alignment.
We experiment with various incorrect target labels within the classes \{1, 3, 5, 7, 9\} of CIFAR-10 and report the average results in Table~\ref{tab:cifar10_backdoor}.
The results demonstrate that amongst other unlearning methods, our framework is the most effective at both countering the backdoor attacks and maintaining the model performance on clean samples.
In contrast, although FT and NegGrad are both relatively successful in reducing the attack success rate, they compromise the overall model performance.
Figure~\ref{fig:umap_backdoor} visualizes the feature representations of the backdoor attacks, before and after unlearning has taken place.
In this figure, points outlined in black represent the samples that have been poisoned by a backdoor trigger.
Figure~\ref{fig:umap_backdoor}(a) demonstrates that these poisoned samples exhibit a shared semantic in the feature representation space.
Our framework successfully eradicates the information pertaining to backdoor trigger, as shown in Figure~\ref{fig:umap_backdoor}(b), purging the backdoor effects from the model.

\paragraph{Training stability}

To verify the stability of our framework during training, we evaluate LP($\mathcal{D}_r$) and LP($\mathcal{D}_f$) as training progresses, and compare these results with those from NegGrad~\cite{thudi2022unrolling}.
Figure~\ref{fig:abl_stability} shows that the model unlearning with our framework quickly converges to the desired performance in terms of LP($\mathcal{D}_r$) and LP($\mathcal{D}_f$), and maintains this performance even with excess training.
On the other hand, the model unlearned with NegGrad experiences quick and sudden changes in LP($\mathcal{D}_r$) and LP($\mathcal{D}_f$), diverging to suboptimal performance with more iterations. 
Thus, the timing of early stopping is imperative to obtain an unlearned model with reasonable performance, making it more difficult to use NegGrad in practical scenarios.


\section{Conclusion}
\label{sec:conclusion}

We presented a novel machine unlearning framework by leveraging feature representations.
We began by developing a novel metric, dubbed as dimensional alignment, which analyzes the unlearning in latent feature spaces by measuring the alignment between eigenspaces of the forget and retain sets. 
This metric serves as both a robust analytical tool and a powerful objective for guiding the unlearning process. 
Our holistic unlearning framework integrates dimensional alignment, self-distillation, and an alternate training scheme to facilitate effective and stable unlearning.
Finally, we highlighted the limitations of established evaluation metrics for machine unlearning and introduced new feature-level evaluation metrics that more accurately reflect the goals of machine unlearning.
We believe these contributions advance our understanding and assessment of unlearning algorithms, moving the field toward more reliable, effective, and transparent unlearning practices in machine learning systems.

\paragraph{Acknowledgements}
This work was partly supported by Samsung Advanced Institute of Technology (SAIT), and by the National Research Foundation of Korea grant [No.2022R1A2C3012210] and the Institute of Information communications Technology Planning \& Evaluation (IITP) grants [No.RS-2022-II220959, No.RS-2021-II211343, No.RS-2021-II212068], funded by the Korean government (MSIT).

{\small
\bibliographystyle{ieee_fullname}
\bibliography{egbib}

\begin{thebibliography}{10}\itemsep=-1pt

\bibitem{alain2016understanding}
Guillaume Alain and Yoshua Bengio.
\newblock Understanding intermediate layers using linear classifier probes.
\newblock {\em arXiv preprint arXiv:1610.01644}, 2016.

\bibitem{bourtoule2021machine}
Lucas Bourtoule, Varun Chandrasekaran, Christopher~A Choquette-Choo, Hengrui
  Jia, Adelin Travers, Baiwu Zhang, David Lie, and Nicolas Papernot.
\newblock Machine unlearning.
\newblock In {\em IEEE SP}, 2021.

\bibitem{cao2015towards}
Yinzhi Cao and Junfeng Yang.
\newblock Towards making systems forget with machine unlearning.
\newblock In {\em IEEE SP}, 2015.

\bibitem{caron2021emerging}
Mathilde Caron, Hugo Touvron, Ishan Misra, Rafal J{\'o}zefowicz, Armand Joulin,
  Piotr Bojanowski, and Matthijs Douze.
\newblock Emerging properties in self-supervised vision transformers.
\newblock In {\em ICCV}, 2021.

\bibitem{cha2024learning}
Sungmin Cha, Sungjun Cho, Dasol Hwang, Honglak Lee, Taesup Moon, and Moontae
  Lee.
\newblock Learning to unlearn: Instance-wise unlearning for pre-trained
  classifiers.
\newblock In {\em AAAI}, 2024.

\bibitem{chundawat2023can}
Vikram~S Chundawat, Ayush~K Tarun, Murari Mandal, and Mohan Kankanhalli.
\newblock Can bad teaching induce forgetting? unlearning in deep networks using
  an incompetent teacher.
\newblock In {\em AAAI}, 2023.

\bibitem{dwork2006differential}
Cynthia Dwork.
\newblock Differential privacy.
\newblock In {\em ICALP}, 2006.

\bibitem{foster2024fast}
Jack Foster, Stefan Schoepf, and Alexandra Brintrup.
\newblock Fast machine unlearning without retraining through selective synaptic
  dampening.
\newblock In {\em AAAI}, 2024.

\bibitem{goel2022evaluating}
Shashwat Goel, Ameya Prabhu, and Ponnurangam Kumaraguru.
\newblock Evaluating inexact unlearning requires revisiting forgetting.
\newblock {\em CoRR abs/2201.06640}, 2022.

\bibitem{golatkar2020eternal}
Aditya Golatkar, Alessandro Achille, and Stefano Soatto.
\newblock Eternal sunshine of the spotless net: Selective forgetting in deep
  networks.
\newblock In {\em CVPR}, 2020.

\bibitem{golatkar2020forgetting}
Aditya Golatkar, Alessandro Achille, and Stefano Soatto.
\newblock Forgetting outside the box: Scrubbing deep networks of information
  accessible from input-output observations.
\newblock In {\em ECCV}, 2020.

\bibitem{graves2021amnesiac}
Laura Graves, Vineel Nagisetty, and Vijay Ganesh.
\newblock Amnesiac machine learning.
\newblock In {\em AAAI}, 2021.

\bibitem{gu2017badnets}
Tianyu Gu, Brendan Dolan-Gavitt, and Siddharth Garg.
\newblock Badnets: Identifying vulnerabilities in the machine learning model
  supply chain.
\newblock {\em arXiv preprint arXiv:1708.06733}, 2017.

\bibitem{guo2019certified}
Chuan Guo, Tom Goldstein, Awni Hannun, and Laurens Van Der~Maaten.
\newblock Certified data removal from machine learning models.
\newblock In {\em ICML}, 2020.

\bibitem{he2022masked}
Kaiming He, Xinlei Chen, Saining Xie, Yanghao Li, Piotr Doll{\'a}r, and Ross
  Girshick.
\newblock Masked autoencoders are scalable vision learners.
\newblock In {\em CVPR}, 2022.

\bibitem{he2016deep}
Kaiming He, Xiangyu Zhang, Shaoqing Ren, and Jian Sun.
\newblock {Deep Residual Learning for Image Recognition}.
\newblock In {\em CVPR}, 2016.

\bibitem{hinton2015distilling}
Geoffrey Hinton, Oriol Vinyals, and Jeff Dean.
\newblock Distilling the knowledge in a neural network.
\newblock In {\em NeurIPS}, 2014.

\bibitem{jacot2018neural}
Arthur Jacot, Franck Gabriel, and Cl{\'e}ment Hongler.
\newblock Neural tangent kernel: Convergence and generalization in neural
  networks.
\newblock In {\em NIPS}, 2018.

\bibitem{kim2023stability}
Dongwan Kim and Bohyung Han.
\newblock On the stability-plasticity dilemma of class-incremental learning.
\newblock In {\em CVPR}, 2023.

\bibitem{koh2017understanding}
Pang~Wei Koh and Percy Liang.
\newblock Understanding black-box predictions via influence functions.
\newblock In {\em ICML}, 2017.

\bibitem{krizhevsky2009learning}
Alex Krizhevsky, Geoffrey Hinton, et~al.
\newblock Learning multiple layers of features from tiny images.
\newblock 2009.

\bibitem{kurmanji2024towards}
Meghdad Kurmanji, Peter Triantafillou, Jamie Hayes, and Eleni Triantafillou.
\newblock Towards unbounded machine unlearning.
\newblock In {\em NeurIPS}, 2023.

\bibitem{le2015tiny}
Ya Le and Xuan Yang.
\newblock Tiny imagenet visual recognition challenge.
\newblock {\em CS 231N}, 7(7):3, 2015.

\bibitem{liu2024model}
Jiancheng Liu, Parikshit Ram, Yuguang Yao, Gaowen Liu, Yang Liu, PRANAY SHARMA,
  Sijia Liu, et~al.
\newblock Model sparsity can simplify machine unlearning.
\newblock In {\em NeurIPS}, 2023.

\bibitem{ma2021sanity}
Xiaolong Ma, Geng Yuan, Xuan Shen, Tianlong Chen, Xuxi Chen, Xiaohan Chen, Ning
  Liu, Minghai Qin, Sijia Liu, Zhangyang Wang, et~al.
\newblock Sanity checks for lottery tickets: Does your winning ticket really
  win the jackpot?
\newblock In {\em NeurIPS}, 2021.

\bibitem{mcinnes2018umap}
Leland McInnes, John Healy, and James Melville.
\newblock Umap: Uniform manifold approximation and projection for dimension
  reduction.
\newblock {\em arXiv preprint arXiv:1802.03426}, 2018.

\bibitem{neel2021descent}
Seth Neel, Aaron Roth, and Saeed Sharifi-Malvajerdi.
\newblock Descent-to-delete: Gradient-based methods for machine unlearning.
\newblock In {\em Algorithmic Learning Theory}, pages 931--962. PMLR, 2021.

\bibitem{paszke2019pytorch}
Adam Paszke, Sam Gross, Francisco Massa, Adam Lerer, James Bradbury, Gregory
  Chanan, Trevor Killeen, Zeming Lin, Natalia Gimelshein, Luca Antiga, et~al.
\newblock Pytorch: An imperative style, high-performance deep learning library.
\newblock In {\em NeurIPS}, 2019.

\bibitem{pereyra2017regularizing}
Gabriel Pereyra, George Tucker, Jan Chorowski, {\L}ukasz Kaiser, and Geoffrey
  Hinton.
\newblock Regularizing neural networks by penalizing confident output
  distributions.
\newblock {\em arXiv preprint arXiv:1701.06548}, 2017.

\bibitem{roy2007effective}
Olivier Roy and Martin Vetterli.
\newblock The effective rank: A measure of effective dimensionality.
\newblock In {\em EUSIPCO}, 2007.

\bibitem{shokri2017membership}
Reza Shokri, Marco Stronati, Congzheng Song, and Vitaly Shmatikov.
\newblock Membership inference attacks against machine learning models.
\newblock In {\em IEEE SP}.

\bibitem{song2019privacy}
Liwei Song, Reza Shokri, and Prateek Mittal.
\newblock Privacy risks of securing machine learning models against adversarial
  examples.
\newblock In {\em ACM SIGSAC}, 2019.

\bibitem{thudi2022unrolling}
Anvith Thudi, Gabriel Deza, Varun Chandrasekaran, and Nicolas Papernot.
\newblock Unrolling sgd: Understanding factors influencing machine unlearning.
\newblock In {\em EuroS\&P}, 2022.

\bibitem{warnecke2021machine}
Alexander Warnecke, Lukas Pirch, Christian Wressnegger, and Konrad Rieck.
\newblock Machine unlearning of features and labels.
\newblock {\em arXiv preprint arXiv:2108.11577}, 2021.

\bibitem{wu2018group}
Yuxin Wu and Kaiming He.
\newblock Group normalization.
\newblock In {\em ECCV}, 2018.

\bibitem{yeom2018privacy}
Samuel Yeom, Irene Giacomelli, Matt Fredrikson, and Somesh Jha.
\newblock Privacy risk in machine learning: Analyzing the connection to
  overfitting.
\newblock In {\em IEEE CSF}, 2018.

\bibitem{zhang2021understanding}
Chiyuan Zhang, Samy Bengio, Moritz Hardt, Benjamin Recht, and Oriol Vinyals.
\newblock Understanding deep learning (still) requires rethinking
  generalization.
\newblock {\em Communications of the ACM}, 64(3):107--115, 2021.

\bibitem{zhang2022prompt}
Zijie Zhang, Yang Zhou, Xin Zhao, Tianshi Che, and Lingjuan Lyu.
\newblock Prompt certified machine unlearning with randomized gradient
  smoothing and quantization.
\newblock In {\em NeurIPS}, 2022.

\end{thebibliography}
}


\setcounter{section}{0}
\setcounter{table}{0}
\setcounter{figure}{0}

\renewcommand\thesection{\Alph{section}}
\renewcommand\thetable{\Alph{table}}
\renewcommand\thefigure{\Alph{figure}}

\newpage

\onecolumn


\section{Related Works}
\label{sec:related}

Machine unlearning approaches are designed to expunge information pertaining to a particular subset of training data from the model weights, while maintaining the model performance on the rest of the data.
The concept of machine unlearning was first introduced in~\cite{cao2015towards} as an efficient forgetting algorithm tailored for statistical query learning.
Bourtoule~\etal~\cite{bourtoule2021machine} proposed a framework that shards data into multiple models, enabling precise unlearning of specific data segments. 
This method ensures complete forgetting but incurs significant storage costs due to the need to maintain multiple models or gradients.
In the context of model interpretability, Koh~\etal~\cite{koh2017understanding} provided a Hessian-based method for estimating the influence of a training point on the model predictions. 
Guo~\etal~\cite{guo2019certified} introduced $\epsilon$-certified removal, which applied differential privacy~\cite{dwork2006differential} to certify the data removal process, and proposed a method for removing information from model weights in convex problems using Newton's method.
Neel~\etal~\cite{neel2021descent} proposed a gradient descent-based method for data deletion in convex settings, providing theoretical guarantees for multiple forgetting requests.
Although these approaches have been proven effective, they are not fully suitable for deep neural network due to its non-convex nature.

Recently, there have been numerous attempts to address machine unlearning in deep neural networks.
Golatkar~\etal~\cite{golatkar2020eternal,golatkar2020forgetting} took an information-theoretic approach to eliminate data-specific information from weights, leveraging the Neural Tangent Kernel (NTK) theory~\cite{jacot2018neural}.
Fisher forgetting~\cite{golatkar2020forgetting} utilized the Fisher information matrix to identify the optimal noise level required to effectively eliminate the influence of samples designated for unlearning.
Liu~\etal~\cite{liu2024model} presented that increasing model sparsity can boost effective unlearning, and proposed a unlearning framework that utilizes pruning methods~\cite{ma2021sanity} on top of existing unlearning approaches.
Chundawat~\etal~\cite{chundawat2023can} used a teacher-student distillation framework, where the student model selectively receives knowledge from both effective and ineffective teachers, facilitating targeted forgetting.
Similarly, Kurmanji~\etal~\cite{kurmanji2024towards} employed a teacher-student network but simplify the approach by using only a single teacher.
Our self-distillation loss shares some similarities with these distillation-based approaches, but it offers clear advantages by targeting an equilibrium, resulting in more stable training.
Unlike previous unlearning works, our primary focus is on latent feature representation space, aimed at effectively mitigating the information leakage problem associated with machine unlearning.

On the other hand, several works have proposed modifications to the original model training to make the resulting model more amenable to unlearning.
Thudi~\etal~\cite{thudi2022unrolling} introduced a regularizer to reduce the verification error, which approximates the distance between the unlearned model and a retrained model, aiming to facilitate easier unlearning in the future. 
Zhang~\etal~\cite{zhang2022prompt} presented a training process that quantizes gradients and applies randomized smoothing, which is designed to make unlearning unnecessary in the future and comes with certifications under some conditions. 
However, these approaches assumes that the deletion request does not cause significant changes in data distribution, which is not applicable to practical scenarios such as class unlearning.

\section{Implementation details}
\label{supp:implementation}

\paragraph{NMI}
To calculate the normalized mutual information (NMI), we initially perform $k$-means clustering on dataset $\mathcal{D}$ based on feature representations, with $k$ equal to the number of classes, $Y$.
Let $\mathbf{K} \in \{1, ..., Y\}^{|\mathcal{D}|}$ represent the cluster assignments for $\mathcal{D}$, and $\mathbf{X} \in \{0, 1\}^{|\mathcal{D}|}$ indicate whether each sample belongs to the forget set.
NMI is then computed using the formula $\frac{I(\mathbf{K}, \mathbf{X})}{\min(H(\mathbf{K}), H(\mathbf{X}))}$, where $I(\cdot, \cdot)$ is the mutual information and $H(\cdot)$ denotes the entropy.

\paragraph{F1 score}
To measure the F1 score, we utilize the same $k$-means clustering.
We calculate the recall and precision for each cluster regarding $\mathcal{D}_f$.
Precision is defined as the proportion of cluster examples that belong to $\mathcal{D}_f$, while recall is the proportion of $\mathcal{D}_f$ assigned to the cluster.
The F1 score is computed by the harmonic mean of recall and precision.
We report the final F1 score for the cluster that yields the highest value, indicating the most relevant cluster to $\mathcal{D}_f$.

\paragraph{Membership inference attack (MIA) success rate}
Following prior work~\cite{liu2024model}, we employ a confidence-based MIA predictor.
Given the unlearned model, $\theta_u$, and the datasets, $\mathcal{D}_f$, $\mathcal{D}_r$, and $\mathcal{D}_\text{test}$, we first calculate the confidence, denoted as $q(\cdot)$, for each example in the datasets.
Then, we train a logistic regression model, $h(\cdot)$, using $\mathcal{D}_r$ and $\mathcal{D}_\text{test}$, which aims to predict $h(q(x)) = 1$ for $x \in \mathcal{D}_r$ and $h(q(x)) = 0$ for $x \in \mathcal{D}_\text{test}$.
We measure the MIA success rate by averaging $h(q(x))$ for all $x \in \mathcal{D}_\text{f}$, where the lower values indicate successful unlearning.

\paragraph{Linear probing}
Given the target model $\theta$, linear probing protocol involves training a new linear classifier on top of its frozen feature extractor.
For evaluating LP($\mathcal{D}_r$), the linear classifier is trained with $ \mathcal{D}_r $, and we report the performance on the $\mathcal{D}_r$ to focus on the discriminability of the retain samples.
To measure LP($\mathcal{D}_f$), we train a linear classifier using $\mathcal{D} = \mathcal{D}_r \cup \mathcal{D}_f$, and report the performance on $\mathcal{D}_f$, which is for evaluating the identifiability of $\mathcal{D}_f$. 

\paragraph{Hyperparameters}
We tune the learning rate for all compared approaches within $\{0.1, 0.01, 10^{-3}, 10^{-4}\}$, except for the NegGrad, for which we use $\{10^{-4}, 10^{-5}\}$.
For EU-$k$ and CF-$k$, we follow the same $k$ with prior work~\cite{goel2022evaluating}, updating the conv4 and fc layers of ResNet while keeping the other layers frozen.
For SCRUB, we follow the original paper's code implementation with $\alpha=0.001$ and $\gamma=0.99$.
For Fisher forgetting, we use the code implementation provided in~\cite{golatkar2020forgetting}.
We set 200 training iterations for our framework.

\section{Additional experimental results}

\subsection{Results on existing evaluation metrics}

To provide a comprehensive view, we evaluate the unlearning algorithms with existing measurements, including Acc($\mathcal{D}_f$), Acc($\mathcal{D}_r$), and the MIA score.
Table~\ref{tab:cifar10_class_full},~\ref{tab:cifar100_class_full}, and~\ref{tab:tiny_class_full} present the overall experimental results.


\begin{table}[t]
\centering
\caption{Unlearning results on the CIFAR-10 dataset averaging over five different configurations.}
\label{tab:cifar10_class_full}
\vspace{3mm}
\setlength\tabcolsep{8pt}
\scalebox{0.9}{
\begin{tabular}{lccccc:ccc}
\toprule
Method & DA & LP($\mathcal{D}_f$) & LP($\mathcal{D}_r$) & F1 & NMI & Acc($\mathcal{D}_f$) & Acc($\mathcal{D}_r$) & MIA \\
\midrule
Original & 0.34 & 92.9 & 92.5 & 0.99 & 0.96 & 92.9 & 92.0 & 0.91 \\
Retrained & 0.79 & 65.4 & 92.1 & 0.54 & 0.31 & ~~0.0 & 92.2 & 0.37 \\
\midrule
FT & 0.51 & 88.3 & {92.8} & 0.80 & 0.65 & 40.2 & 92.8 & 0.21 \\
FT (classifier only) & 0.34 & 92.9 & {92.5} & 0.99 & 0.96 & ~~0.0 & 92.8 & 0.00 \\
NegGrad & 0.55 & {66.8} & 90.2 & {0.51} & {0.23} & ~~3.7 & 85.2 & 0.63 \\
Fisher & 0.37 & 88.5 & 90.2 & 0.97 & 0.89 & ~~8.4 & 88.6 & 0.01 \\
SCRUB & 0.41 & 74.7 & 92.0 & 0.76 & 0.59 & 50.2 & 91.8 & 0.46 \\
EU-$k$ & 0.73 & 68.1 & 90.7 & 0.73 & 0.46 & ~~0.0 & 90.9 & 0.19 \\
CF-$k$ & 0.60 & 81.3 & \textbf{92.1} & 0.66 & 0.43 & 13.7 & 92.1 & 0.15 \\
MUDA (Ours) & \textbf{0.82} & \textbf{66.4} & {92.3} & \textbf{0.54} & \textbf{0.32} & ~~0.0 & 92.3 & 0.29 \\
\bottomrule
\end{tabular}
}
\end{table}


\begin{table}[t]
\centering
\caption{Unlearning results on the CIFAR-100 dataset averaging over five different configurations.}
\vspace{2mm}
\setlength\tabcolsep{8pt}
\scalebox{0.9}{
\begin{tabular}{lccccc:ccc}
\toprule
Method & DA & LP($\mathcal{D}_f$) & LP($\mathcal{D}_r$) & F1 & NMI & Acc($\mathcal{D}_f$) & Acc($\mathcal{D}_r$) & MIA \\
\midrule
Original & 0.50 & 75.0 & 72.1 & 0.73 & 0.64 & 76.8 & 72.3 & 0.91 \\
Retrained & 0.74 & 50.8 & 71.2 & 0.33 & 0.19 & ~~0.0 & 71.4 & 0.18 \\
\midrule
FT & 0.58 & 70.2 & 70.7 & 0.71 & 0.61 & 44.0 & 71.2 & 0.08 \\
FT (classifier only) & 0.54 & 74.2 & 70.5 & 0.99 & 0.97 & ~~0.0 & 66.3 & 0.01 \\
NegGrad & 0.55 & 52.0 & \textbf{71.3} & 0.73 & 0.60 & 20.0 & 71.2 & 0.15 \\
Fisher & 0.59 & 67.0 & 62.7 & 0.87 & 0.78 & ~~0.0 & 62.3 & 0.06 \\
SCRUB & 0.60 & 60.0 & 70.1 & 0.68 & 0.57 & 32.0 & 70.3 & 0.32 \\
EU-$k$ & 0.70 & 35.2 & 37.1 & 0.26 & 0.14 & ~~0.0 & 28.0 & 0.45 \\
CF-$k$ & 0.54 & 72.6 & 69.4 & 0.93 & 0.90 & 70.8 & 69.5 & 0.81 \\
MUDA (Ours) & \textbf{0.73} & \textbf{37.0} & \textbf{71.1} & \textbf{0.34} & \textbf{0.21} & ~~0.0 & 71.1 & 0.15 \\
\bottomrule
\end{tabular}
}
\label{tab:cifar100_class_full}
\end{table}


\begin{table}[t]
\centering
\caption{Unlearning results on the Tiny-ImageNet dataset.}
\label{tab:tiny_class_full}
\vspace{2mm}
\setlength\tabcolsep{8pt}
\scalebox{0.9}{
\begin{tabular}{lc:cccc:ccc}
\toprule
Method & DA & LP($\mathcal{D}_f$) & LP($\mathcal{D}_r$) & F1 & NMI & Acc($\mathcal{D}_f$) & Acc($\mathcal{D}_r$) & MIA \\
\midrule
Original & 0.59 & 47.6 & 58.0 & 0.96 & 0.92 & 51.2 & 59.1 & 0.89 \\
Retrained & 0.73 & 24.0 & 58.0 & 0.19 & 0.09 & ~~0.0 & 59.1 & 0.14 \\
\midrule
FT & 0.60 & 45.6 & 56.2 & 0.66 & 0.55 & 44.8 & 57.3 & 0.78 \\
FT (classifier only) & 0.61 & 47.6 & \textbf{58.0} & 0.66 & 0.55 & ~~0.0 & 41.2 & 0.18 \\
NegGrad & \textbf{0.74} & {29.6} & 55.1 & {0.22} & {0.12} & ~~0.0 & 51.8 & 0.30 \\
Fisher & 0.66 & 34.8 & 47.0 & 0.39 & 0.25 & ~~0.0 & 43.1 & 0.20 \\
SCRUB & 0.64 & 35.6 & 55.8 & 0.52 & 0.40 & 38.4 & 56.2 & 0.50 \\
EU-$k$ & 0.77 & 12.8 & 19.1 & 0.11 & 0.04 & ~~0.0 & 11.1 & 0.42 \\
CF-$k$ & 0.63 & 40.8 & 52.7 & 0.48 & 0.33 & 31.2 & 51.6 & 0.47 \\
MUDA (Ours) & 0.70 & \textbf{26.0} & {57.4} & \textbf{0.21} & \textbf{0.10} & ~~0.0 & 58.1 & 0.03 \\
\bottomrule
\end{tabular}
}
\end{table}

\section{Discussion}

\subsection{Random sample unlearning}
\label{sec:random_sample_unlearning}
While most existing works have been evaluated under a random sample unlearning scenario, we did not explicitly address this setting.
We argue that if the forget set is randomly drawn from the training set, these random forget samples do not provide meaningful additional information beyond the remaining samples, implying no information needs to be removed.

To clarify this, assume that both $\mathcal{D}_\text{old}$ and $\mathcal{D}_\text{new}$ follow the same distribution as $\mathcal{D}$.
Given an old model trained on $\mathcal{D}_\text{old}$, we consider a incremental learning scenario involving $\mathcal{D}_\text{new}$.
Since $\mathcal{D}_\text{new}$ follows the same distribution as $\mathcal{D}_\text{old}$, it behaves similar as $\mathcal{D}_\text{old}$ and the decision boundary would not change significantly during incremental learning.
As there are no substantial changes caused by $\mathcal{D}_\text{new}$, unlearning $\mathcal{D}_\text{new}$ should rarely impact the model parameters.

Furthermore, if a user requests a random subset of samples to be forgotten, it is unclear whether the request refer to the specific selected samples or the entire (sub)class corresponding to those samples.
Therefore, we limit the unlearning scenario to cases where the forget set contains meaningful semantics, such as a class, subclass, or group.

Note that the experimental results on defending backdoor attack in Section~\ref{sec:analysis} implicitly address random sample unlearning, where the forget set is a random subset of training set. 
The difference is that in each application, the forget samples share a common semantic, \eg, containing a black patch or label noise.

\subsection{Exploiting forget set accuracy and MIA}
\label{supp:metric_limitations}
The forget set accuracy and MIA can be easily exploited with trivial fine-tuning or post-processing techniques, which render them unreliable for adequately evaluating the unlearned model.  
Below we provide examples of such trivial methods that can easily exploit/circumvent each metric.

\paragraph{Forget set accuracy}
Achieving $0\%$ accuracy on the forget set can be accomplished by simply setting the bias value of the corresponding class in the classifier to $-\infty$, ensuring that no samples are predicted for that class. 
This implies that merely matching the accuracy on the forget set does not necessarily indicate successful unlearning.

\paragraph{MIA success rate}
Since MIA leverages model outputs, such as confidence scores or entropy~\cite{song2019privacy,yeom2018privacy}, to infer the presence of a sample in the training dataset, it depends heavily on the model overfitting to the dataset. 
The underlying assumption is that a model will produce more confident predictions for samples it has seen during training compared to unseen data. 
Hence, if models undergo uncertainty calibration via fine-tuning or post-processing techniques, the MIA may significantly overestimate the effectiveness of unlearning.
To empirically support our claim, we fix the feature extractor of $\theta_o$ and only fine-tune its final linear classifier using a calibration loss~\cite{pereyra2017regularizing}.
We observe that simple post-processing calibration with minimal training significantly lowers the MIA score from 0.91 to 0.02 on CIFAR-10, despite no particular efforts to \textit{unlearn}.

\subsection{Limitations}

Our paper, like previous studies, shares a common limitation: the lack of a theoretical guarantee regarding unlearning.
Nonetheless, our framework introduces a unique approach by focusing on feature representation, which supplements previous research efforts by offering a novel and thorough analysis of machine unlearning.
Additionally, our dimensional alignment loss requires some amount of retain samples, but we have shown that only a minimal number of these samples are necessary to attain effective unlearning performance.

\subsection{Broader impact}

By enabling the effective removal of data from machine learning models without requiring complete retraining, machine unlearning helps organizations comply with privacy laws such as GDPR and the CCPA, which mandate the right to be forgotten.
This is crucial in situations where users withdraw their consent for data use or when data must be deleted for legal reasons. 
Moreover, machine unlearning reduces the risks associated with data breaches, as it ensures sensitive information can be dynamically and reliably erased from models, thus limiting potential misuse.
Additionally, this research can lead to more sustainable AI practices by reducing the computational and environmental costs associated with retraining models from scratch.
This leads to more ethical AI systems by promoting transparency, user trust, and the responsible use of data.

\end{document}